\documentclass[sigconf]{acmart}

\acmConference[HotCarbon'24]{3rd Workshop on Sustainable Computer
Systems}{July 9, 2024}{Santa Cruz, CA}
\acmDOI{}
\acmISBN{}
\setcopyright{rightsretained}

\usepackage[normalem]{ulem}
\usepackage{enumitem}
\usepackage{rotating}
\usepackage{hyperref}
\usepackage{booktabs}
\usepackage{balance}
\usepackage{graphicx}
\usepackage{wrapfig}
\usepackage{algorithm}
\usepackage[noend]{algpseudocode}
\usepackage{amsmath}
\usepackage{fancyhdr}
\usepackage{bbm}
\usepackage{multirow}
\usepackage{xcolor}
\usepackage{soul}
\usepackage{tikz}
\usepackage{cleveref}
\usepackage{url}
\usepackage{mathtools}
\usepackage{pifont}
\usepackage{makecell}
\usepackage{subcaption}
\usepackage{xspace}
\usepackage[skins]{tcolorbox}
\usepackage{array}
\newcommand{\PreserveBackslash}[1]{\let\temp=\\#1\let\\=\temp}
\newcolumntype{C}[1]{>{\PreserveBackslash\centering}p{#1}}
\newcolumntype{R}[1]{>{\PreserveBackslash\raggedleft}p{#1}}
\newcolumntype{L}[1]{>{\PreserveBackslash\raggedright}p{#1}}

\creflabelformat{equation}{#2\textup{#1}#3}

\newlength{\myeqskip}  
\AtBeginDocument{%
    \setlength\abovedisplayskip{\myeqskip}%
    \setlength\belowdisplayskip{\myeqskip}%
    \setlength\abovedisplayshortskip{\myeqskip-\baselineskip}%
    \setlength\belowdisplayshortskip{\myeqskip}}

\newcommand{\todo}[1]{{\color{red}[TODO: #1]}}

\setlist{noitemsep, leftmargin=*, topsep=0pt, partopsep=0pt}

\newcommand{\tf}{T4\xspace}
\newcommand{\ada}{RTX6000~Ada\xspace}

\usepackage{tikz}
\usetikzlibrary{external}
\usepackage{pgfplots}
\usepackage{pgfplotstable}
\usetikzlibrary{pgfplots.groupplots}
\usetikzlibrary{arrows}
\usetikzlibrary{patterns}
\usetikzlibrary{positioning}
\usetikzlibrary{decorations.pathreplacing}
\usetikzlibrary{shapes.arrows}
\usetikzlibrary{shapes.geometric,shapes.misc}
\usetikzlibrary{pgfplots.groupplots}
\pgfplotsset{compat=newest}

\tcbset{
  takeaway/.style={ enhanced,width=\hsize,left=0pt,right=0pt,top=0pt,bottom=0pt,colback=black!20!green!8!white,boxrule=1pt,colframe=black!30!green!30!white
  },
}
\newcounter{takeawaycounter}
\newcommand{\takeawaybox}[1]{
    \stepcounter{takeawaycounter}
    \begin{tcolorbox}[takeaway]
    \textbf{Takeaway~\arabic{takeawaycounter}:}#1
    \end{tcolorbox}
}

\tcbset{
  question/.style={ enhanced,width=\hsize,left=0pt,right=0pt,top=0pt,bottom=0pt,colback=black!20!red!8!white,boxrule=1pt,colframe=black!30!red!30!white
  },
}
\newcounter{questioncounter}

\Crefname{figure}{Figure}{Figures}

\begin{document}

\title{Towards Sustainable Large Language Model Serving}



\author{Sophia Nguyen}
\affiliation{%
   \institution{University of Waterloo}
   \city{Waterloo}
   \state{ON}
   \country{CAN}}
\email{s62nguye@uwaterloo.ca}
\authornote{Both authors contributed equally and are listed in alphabetical order by last name.}

\author{Beihao Zhou}
\affiliation{%
   \institution{University of Waterloo}
   \city{Waterloo}
   \state{ON}
   \country{CAN}}
\email{b72zhou@uwaterloo.ca}
\authornotemark[1]

\author{Yi Ding}
\affiliation{%
   \institution{Purdue University}
   \city{West Lafayette}
   \state{IN}
   \country{USA}}
\email{yiding@purdue.edu}

\author{Sihang Liu}
\affiliation{%
   \institution{University of Waterloo}
   \city{Waterloo}
   \state{ON}
   \country{CAN}}
\email{sihangliu@uwaterloo.ca}

\begin{abstract}

In this work, we study LLMs from a carbon emission perspective, addressing both operational and embodied emissions, and paving the way for sustainable LLM serving. 
We characterize the performance and energy of LLaMA with 1B, 3B, and 7B parameters using two Nvidia GPU types, a latest-generation \ada and an older-generation \tf.
We analytically model operational carbon emissions based on energy consumption and carbon intensities from three grid regions --- each representing a different energy source mix, and embodied carbon emissions based on chip area and memory size.
%
%
Our characterization and modeling provide us with an in-depth understanding of the performance, energy, and carbon emissions of LLM serving.
Our findings highlight the potential for optimizing sustainable LLM serving systems by considering both operational and embodied carbon emissions simultaneously.

\end{abstract}

\maketitle

\section{Introduction}\label{sec:intro}
Large language models (LLMs) have revolutionized numerous industries, including search engines, natural language processing, and programming~\cite{vu2023freshllms,shen2024hugginggpt,liu2024your}. Their widespread adoption has significantly increased energy demands. The cycle of LLM deployment includes training and inference/serving. Recent studies reveal that LLM serving now exceeds training in energy consumption, leading to substantial environmental impacts, notably in carbon emissions~\cite{chien2023reducing}. The carbon emissions stemming from the energy consumption of these applications are referred to as \emph{operational carbon emissions} -- typically quantified as carbon dioxide equivalent (CO\textsubscript{2}eq). For instance, serving one prompt in ChatGPT generates more than 4 grams of CO\textsubscript{2}eq \cite{chatgptcarbon2023} -- over 20 times the carbon emission of a web search query \cite{whyyourinternet2020}. 

The development of LLMs not only demands significant energy but also requires substantial computing hardware resources. Training and serving LLMs require powerful GPUs and machine learning (ML) accelerators, such as NVIDIA HGX~\cite{hgx} and Google TPU~\cite{google_tpu}. These devices are typically equipped with large processor chips manufactured with advanced feature sizes (e.g., 5 nm CMOS) and high-bandwidth, high-capacity onboard memory, enabling efficient execution of LLMs. Nevertheless, the manufacturing process of the hardware entails substantial carbon emissions, known as \emph{embodied carbon emissions}. These emissions are particularly prominent in the case of the latest high-performance devices, as demonstrated in prior studies~\cite{gupta2022act,li2023sustainable,zhang2023embodied}.


To mitigate the environmental impact of LLM serving, a thorough understanding of its carbon emissions at a granular level, such as per-token level across various hardware platforms, is essential. However, this aspect has been under-explored. Some studies have only focused on performance metrics. For example, SplitWise~\cite{patel2024splitwise} divides LLM serving into two phases---prefill and decode---and executes them on different GPUs; PowerInfer reduces LLMs' need for GPU onboard memory to better adapt LLMs to consumer-grade GPUs \cite{song2023powerinfer}. Another line of work examines the carbon emissions of traditional ML and web service applications~\cite{patterson2022carbon,luccioni2023counting,wang2023peeling}. However, LLMs differ significantly from these applications due to their highly compute and memory-intensive nature. Most recent work on carbon modeling, such as ACT~\cite{gupta2022act}, LLMCarbon~\cite{faiz2024llmcarbon}, and carbon accounting for BLOOM~\cite{luccioni2024bloomcarbon}, focus on end-to-end, high-level counting without profiling LLMs on real-world hardware platforms. Therefore, there is a gap in understanding the environmental impacts of LLM serving at a granular level.

To bridge this gap, we characterize LLMs through low-level profiling and modeling to understand the environmental impact of LLM serving. We examine the carbon emissions of three parameter sizes of Meta's LLaMA model \cite{touvron2023llama} -- 1B, 3B, and 7B -- on two different types of GPUs: a latest-generation Nvidia \ada from 2023 and an older-generation Nvidia \tf released back in 2018. 
We first characterize the performance and energy consumption of different \emph{configurations} (i.e., model parameter size and batch size), including both per-query and per-token measurement in the prefill and decode phases of LLM serving execution (\Cref{sec:profiling}). 
Then, we model their operational and embodied carbon emissions, respectively (\Cref{sec:carbon}). For operational carbon, we analyze the emissions by considering three regions with different carbon intensities (CIs)~\cite{carbonintensity}, expressed as CO\textsubscript{2}eq emissions per kWh of electricity consumption. For embodied carbon, we analytically model emissions based on chip area and memory size using existing modeling tools \cite{gupta2022act}. We conclude with a discussion of open questions and future directions based on our characterization and modeling findings (\Cref{sec:future}).
%
We summarize our key findings as follows:
\begin{itemize}
    \item The older and slower \tf has higher energy efficiency compared to the newer and faster \ada when processing less compute-intensive requests (e.g., batch size of 1), especially in the decode phase of LLM serving.
    This finding indicates that older hardware can be beneficial in specific configurations. 
    \item A configuration (i.e., model parameter size and batch size) that achieves the highest throughput on a particular GPU type does not necessarily yield the highest energy efficiency, highlighting the complex tradeoffs in LLM serving design space exploration.
    \item By considering both embodied and operational carbon, an energy-efficient configuration may not necessarily minimize carbon emissions. Instead, factors such as model parameters, batch size, and GPU platform collectively influence carbon emissions.
    \item 
    Strategically using older GPUs like \tf could effectively reduce total carbon emissions by amortizing the embodied carbon emissions of GPUs over time.    

\end{itemize}

\section{LLM Characterization}\label{sec:profiling}
In this section, we characterize the performance and energy consumption of LLM serving.

\subsection{Methodology} \label{subsec:perf_meth}
This section describes the methodology of the characterizations. 

\begin{table}[t]
    \centering
    \small
    \caption{Specifications of GPUs in this work.}
    \label{tab:gpus}
    \begin{tabular}{lll}
    \toprule
    GPU Type    &  \ada & \tf \\
    \midrule
    Architecture & Ada Lovelace & Tesla \\
    Chip Size &  608.4 mm\textsuperscript{2} & 545 mm\textsuperscript{2} \\
    Technology Node & 5 nm & 12 nm \\
    Memory    & 48 GB  & 16 GB\\
    Thermal Design Power     & 300 W & 70 W \\
    Year    & 2023 & 2018 \\
    \midrule
    Embodied Carbon   &  26.6 kg  & 10.3 kg \\
    \bottomrule
    \end{tabular}
\end{table}

\textbf{Model and dataset.}
We characterize LLM serving using the widely-used LLaMA model \cite{touvron2023llama} with 1B, 3B, and 7B parameters.
For all experiments, we evaluate prompts from the Alpaca dataset \cite{alpaca}. 
While large-scale LLMs (e.g., those with hundreds of billions of parameters) are powerful, recent work has shown smaller models (e.g., 1B parameters) can achieve high accuracy and are especially valuable in many scenarios, e.g., speculative decoding~\cite{leviathan2023fast}, fine-tuned for specific tasks~\cite{juneja2024small}, working in cooperation with large models \cite{xu2023small}, and retrieval-augmented generation (RAG)~\cite{raft2024}.

\textbf{GPU platforms.}
We evaluate LLMs on two GPU platforms: \ada and \tf, as listed in \Cref{tab:gpus}.
\ada is based on Nvidia's newer Ada Lovelace architecture that became available in 2023.
It has a large 48 GB onboard memory and is manufactured with the advanced 5~nm node~\cite{ada6000techpowerup}. 
\tf is based on an older Nvidia Tesla architecture, equipped with a smaller 16 GB onboard memory that was released in 2018 and manufactured with an older 12 nm node \cite{t4techpowerup}. 
\ada has a much higher 300 W thermal design power (TDP) as compared to \tf's 70 W.

\textbf{Measurement.}
As the output length varies by prompts and model sizes, we time LLM execution for 150 tokens and consider only prompts generating more than 150 tokens when comparing end-to-end latency and energy consumption. As LLMs typically run on GPUs and mainly utilize GPU resources, this study focuses on GPU power consumption. We use NVML \cite{nvml} to measure the GPU's power every 100 ms. 
The energy consumption ($E_{\rm prompt}$) of a prompt is the product of the average GPU power ($P_{\rm prompt}$) and the execution time of the prompt ($t_{\rm prompt}$):
\begin{equation}
    E_{\rm prompt} = P_{\rm prompt} \cdot t_{\rm prompt}
    \label{equ:eng}
\end{equation}



\begin{figure}[t]
    \centering
    \begin{subfigure}[b]{1\linewidth}
    \includegraphics[width=1\linewidth]{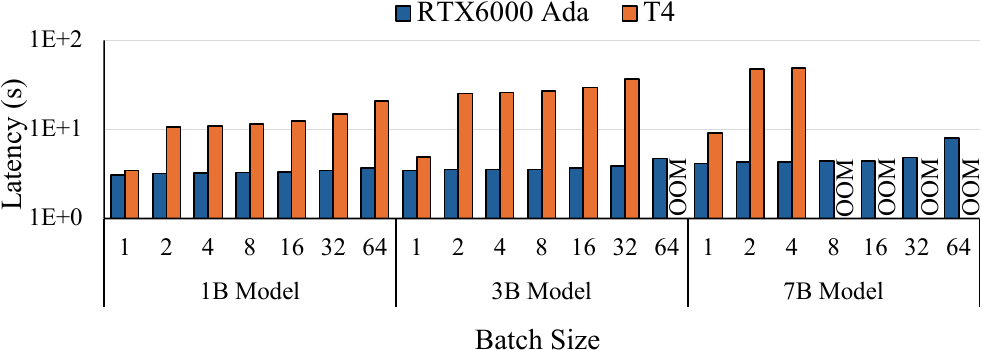}
    \caption{Latency.\vspace{0.1in}}
    \label{fig:lat}
    \Description{}
    \end{subfigure}
    \begin{subfigure}[b]{1\linewidth}
    \includegraphics[width=1\linewidth]{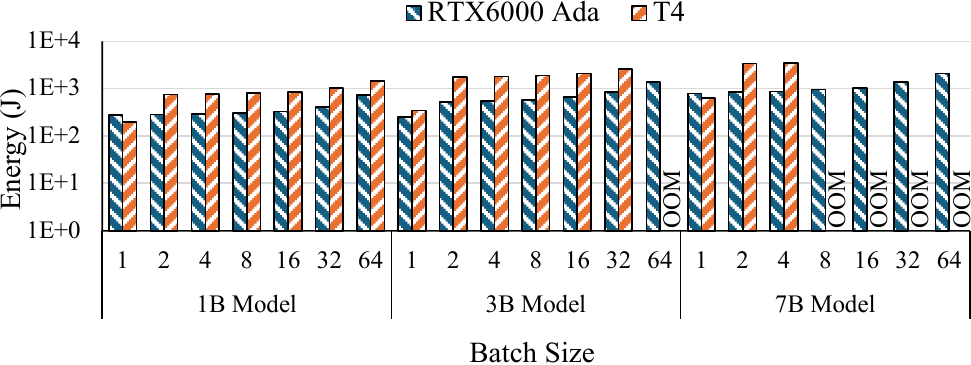}
    \caption{Energy.}
    \label{fig:eng}
    \Description{}
    \end{subfigure}
    \caption{Latency and energy consumption of \ada and \tf under different parameter sizes and batch sizes. ``OOM''=out of memory.
    }
\end{figure}

\subsection{Latency vs. Energy Consumption}
\label{subsec:lat_eng}

We first compare the latency and energy consumption across LLaMA models with different parameter sizes and GPU platforms. 
We report the average power consumption and median latency values for all prompts, during the execution of a specific model/batch size. 

%
\textbf{Latency.}
\Cref{fig:lat} shows the median prompt processing latency~(s) (log-scaled y-axis) across different model sizes from 1B to 3B and batch sizes from 1 to 64 (x-axis).
Across all parameter sizes, the latency is higher in \tf compared to \ada. 
When the batch size is 1, \tf is 1.1$\times$, 1.4$\times$, and 2.2$\times$ slower than \ada when running 1B, 3B, and 7B-parameter models, respectively. 
\tf is more prone to latency increase than \ada as the batch size and parameter size of the model increases, as \tf is an older, lower-tier GPU compared to \ada. 
\tf is up to 11.4$\times$ slower than \ada when running the 7B model with a batch size of 4. 
For sufficiently large model and batch sizes, the 16 GB memory on T4 renders it unsuitable for outputting results, leading to "OOM" (out of memory) indications.

\textbf{Energy consumption.}
\Cref{fig:eng} shows the energy consumption~(J) (log-scaled y-axis), which is calculated from the average power consumption when executing a specific model/batch size (x-axis) and the previous median latency, using \Cref{equ:eng}. 
For both \ada and \tf, the energy consumption increases as the parameter and batch size increase, due to the higher compute intensity and longer execution time.

Surprisingly, despite \ada being a recently released GPU that is better optimized for LLMs,  the energy consumption of \tf is 28 \% and 20 \% lower than \ada when executing a batch size of 1 in the 1B and 7B parameter models, respectively. And, the 3B model shows the energy consumption of \tf is not much higher than \ada (1.4$\times$). 
The main reasons are two-fold. First, a batch size of 1 has a lower computational load. Even \tf is capable of executing such a load efficiently. 
Second, \ada's TDP is more than 4$\times$ higher than \tf. Therefore, under such a light load, \ada is less power-efficient than \tf.

When the batch size increases, \tf consumes more energy than {\ada} (up to 2.9$\times$) -- in these more compute-intensive scenarios, the newer \ada has a significant advantage over the older and less performant \tf. 
Despite the higher TDP, \ada executes the prompts an order of magnitude faster than \tf and thus reduces the per-prompt energy consumption. 

\takeawaybox{
\ada is faster than \tf, regardless of the parameter size or batch size. 
However, when executing a batch size of 1, \tf may have an advantage over \ada, because \tf runs at a much lower power without major degradation in execution time. 
\ada becomes more energy-efficient under larger batch sizes, attributed to its faster processing speed. 
}

\subsection{Prefill and Decode Phases}
\label{subsec:phases}

LLM execution consists of \emph{prefill} and \emph{decode} phases. 
The prefill phase involves processing the prompt, initializing the model state, and then generating the first token.
Afterward, the decode phase generates the rest of the tokens in an auto-regressive fashion, until the termination token (or output token limit) is reached.
%
%
Prior work has shown that these two phases have different performance characteristics \cite{patel2024splitwise, gao2024attentionstore,zhong2024distserve}, where the prefill phase is compute-bounded and more compute-intensive, and the decode phase is memory-bounded and less compute-intensive. 
In this experiment, we study not only the performance of both phases but also their energy consumption. 
Because \tf experiences OOM when running large batch and parameter sizes, we evaluate the LLaMA with 1B parameters.

\begin{figure}[t]
\begin{subfigure}[b]{0.48\linewidth}
    \includegraphics[height=3.5cm]{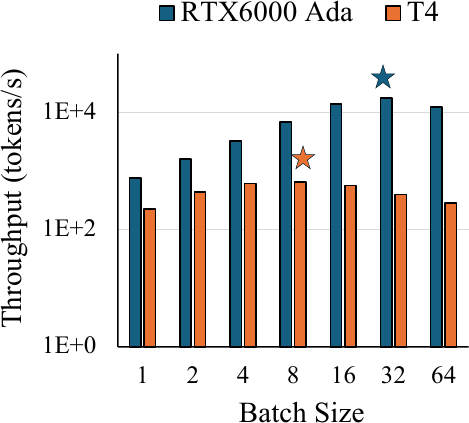}
    \caption{Throughput.}\label{subfig:1b_prefill_throughput}
    \Description{}
\end{subfigure}
\hspace{1pt}
\begin{subfigure}[b]{0.48\linewidth}
    \includegraphics[height=3.5cm]{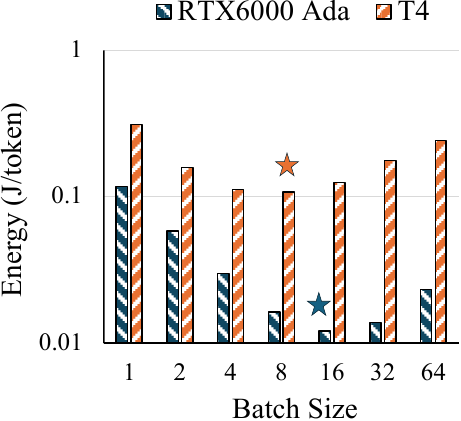}
    \caption{Per-token energy.}\label{subfig:1b_prefill_energy}
    \Description{}
\end{subfigure}
\vspace{-0.2cm}
\caption{Throughput and energy in the \emph{prefill} phase (1B-parameter LLaMA).}
\end{figure}

\begin{figure}[t]
\begin{subfigure}[b]{0.48\linewidth}
    \includegraphics[height=3.5cm]{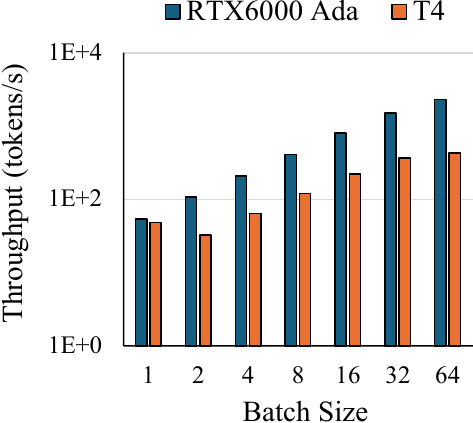}
    \caption{Throughput.}\label{subfig:1b_decode_throughput}
    \Description{}
\end{subfigure}
\hspace{1pt}
\begin{subfigure}[b]{0.48\linewidth}
    \includegraphics[height=3.5cm]{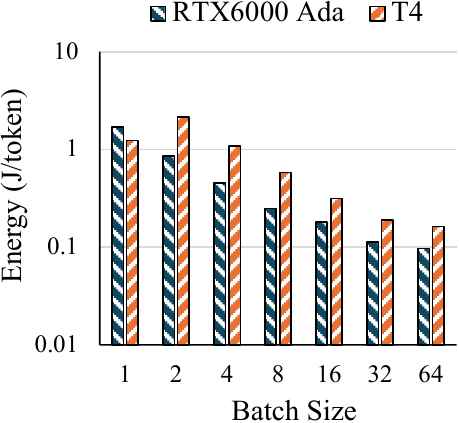}
    \caption{Per-token energy.}\label{subfig:1b_decode_energy}
    \Description{}
\end{subfigure}
\vspace{-0.2cm}
\caption{Throughput and energy in the \emph{decode} phase (1B-parameter LLaMA).}
\end{figure}

\textbf{Prefill Phase.}
We evaluate the prefill phase by measuring its throughput in processing input prompts (tokens/s) and the corresponding per-token energy consumption (J/token).
\Cref{subfig:1b_prefill_throughput} shows the throughput and \Cref{subfig:1b_prefill_energy} energy consumption (log-scaled y-axis) for different batch sizes (x-axis) when running on \ada and \tf.
We observe that \ada is more efficient from both a throughput and energy perspective during the prefill phase. 
This is because the prefill phase is compute-bounded, which prefers more powerful GPUs like \ada. 

We also notice that the throughput reaches the peak when the batch size is 8 on \tf and 32 on \ada. 
Energy-wise, the per-token energy is lowest when batch size is 8 on \tf and 16 on \ada. 
\ada achieves higher throughput and lower per-token energy in a larger batch size than \tf because it can better handle compute-intensive scenarios. 
However, the batch size that achieves the highest throughput is not necessarily the same as which achieves the highest energy efficiency. 

\textbf{Decode Phase.}
We evaluate the decode phase by measuring the throughput (tokens/s) and per-token energy consumption (J/token). 
\Cref{subfig:1b_decode_throughput} and \Cref{subfig:1b_decode_energy} show the throughput and per-token energy (log-scaled y-axis), respectively, in different batch sizes (x-axis). 
Comparing \ada and \tf in the decode phase, \ada always outperforms \tf in throughput but the difference is not as significant as the prefill phase, as the decode phase is memory-bound and less compute-intensive than the prefill phase. 
From an energy perspective, the per-token energy consumption of \tf is lower than \ada when the batch size is 1 --- \tf consumes 27.1\% less energy than \ada but only 9.5\% lower throughput. 
As the batch size increases, \ada becomes more energy-efficient --- up to 5.4$\times$ higher throughput (when the batch size is 64) and 57.5\% lower per-token energy consumption than \tf (when the batch size is 16).
Although older GPUs like \tf are less powerful, their performance is not much worse than \ada while consuming less energy in a batch size of 1, similar to the observations in \Cref{subsec:lat_eng}.
We further observe that the throughput and per-token energy consumption generally improve as the batch size increases, unlike the prefill phase.

\takeawaybox{ Dividing LLM serving into prefill and decode phases reveals more energy optimization opportunities, including distributing them across different GPU platforms. In the prefill phase, the batch size with the highest throughput may not yield the best energy efficiency, indicating an intricate interplay between batch size, performance, and energy consumption. In the decode phase, the optimal tradeoff between energy efficiency and batch size also varies across GPU platforms.

}
\section{Carbon Emission Analysis}\label{sec:carbon}

In this section, we analyze the carbon emissions of LLM serving. 

\subsection{Methodology}
\label{subsec:carbon_meth}

We model and analyze the total carbon emissions for LLM serving, which include operational and embodied carbon emissions.

\textbf{Operational Carbon.} 
We model the operational carbon $C_{\rm op}$ based on the energy consumption and carbon intensity ($CI$) of the grid. The energy consumption of a prompt, $E_{\rm prompt}$, is the energy consumed during the execution time, as we have shown in \Cref{subsec:lat_eng}. Therefore, the operational carbon of a prompt is:
\begin{equation}
C_{\rm prompt,op} = E_{\rm prompt}\cdot CI 
\label{equ:op}
\end{equation}
%

In this work, we study CIs of grids in three regions~\cite{carbonintensity}: Qu\'{e}bec (QC), California Independent System Operator (CISO), and PacifiCorp East (PACE).
\Cref{tab:carbon_intensity} lists their average CIs in 2023, which will be used in the rest of this paper.
We select these regions due to their distinct energy mixes:
QC relies heavily on renewable hydro and wind energy,
CISO incorporates renewable solar energy alongside non-renewable natural gas,
while PACE depends on non-renewable sources like coal and natural gas.
Overall, a higher fraction of renewable energy sources leads to a lower CI.

\textbf{Embodied Carbon.} 
We model the embodied carbon  ($C_{\rm em}$) using the approach in ACT \cite{gupta2022act}, by considering the processor chip areas and memory capacities from the specifications of \ada~\cite{ada6000techpowerup} and \tf \cite{t4techpowerup}.
The total embodied carbon emissions of both GPU types are listed in~\Cref{tab:gpus}, which are close to a prior study on GPU embodied carbon emissions~\cite{li2023sustainable}.
The ACT approach discounts the embodied carbon emissions by the ratio between total execution time ($T$) and GPU's lifetime ($LT$), i.e., ${T}/{LT}\cdot{C_{\rm em}}$.
Therefore, a prompt executed by the GPU that lasts $t_{\rm prompt}$ generates the embodied carbon emission of:
\begin{equation}
C_{\rm prompt,em} = \frac{t_{\rm prompt}}{\mathit{LT}}\cdot{C_{\rm em}}
\label{equ:em}
\end{equation}
In this work, we assume a total GPU lifetime of 5 years -- a typical lifetime of datacenter components \cite{gupta2022act,ostrouchov2020gpulife}.

\textbf{Total Carbon Emissions.}
The total carbon emission consists of embodied and operational carbon emissions.
For a prompt that executes $t_{\rm prompt}$ time, its total carbon emission is:
\begin{equation}
\begin{split}
    C_{\rm prompt} &= C_{\rm prompt,op} + C_{\rm prompt,em} \\
               &= E_{\rm prompt}\cdot CI + \frac{t_{\rm prompt}}{\mathit{LT}}{C_{\rm em}}
    \label{equ:tot}
\end{split}
\end{equation}

\begin{table}
    \setlength{\tabcolsep}{2pt}
    \caption{Carbon intensities (CIs) of three regions in 2023 \cite{carbonintensity}.}
    \label{tab:carbon_intensity}
    \centering
    \small
    \begin{tabular}{L{2cm}L{2cm}L{1.8cm}L{1.8cm}}
        \toprule
        \textbf{Region} & \textbf{QC} & \textbf{CISO} & \textbf{PACE}  \\
        \midrule
        \textbf{State/Province} & QC (Canada) & CA (USA) & WY, UT, AZ, NM, ID (USA)   \\
        \textbf{Main Sources} & Hydro, Wind   & Gas, Solar & Coal, Gas \\ 
        \textbf{CI (g/kWh)}  & \textbf{\textcolor{black!30!green}{31}} & \textbf{\textcolor{black!30!red!40!yellow}{262}} & \textbf{\textcolor{black!15!red}{647}} \\
        \bottomrule
    \end{tabular}
\end{table}

\subsection{Carbon Emissions in Different Regions}
\label{subsec:location_carbon}

\begin{figure}
    \centering
    \includegraphics[width=1\linewidth]{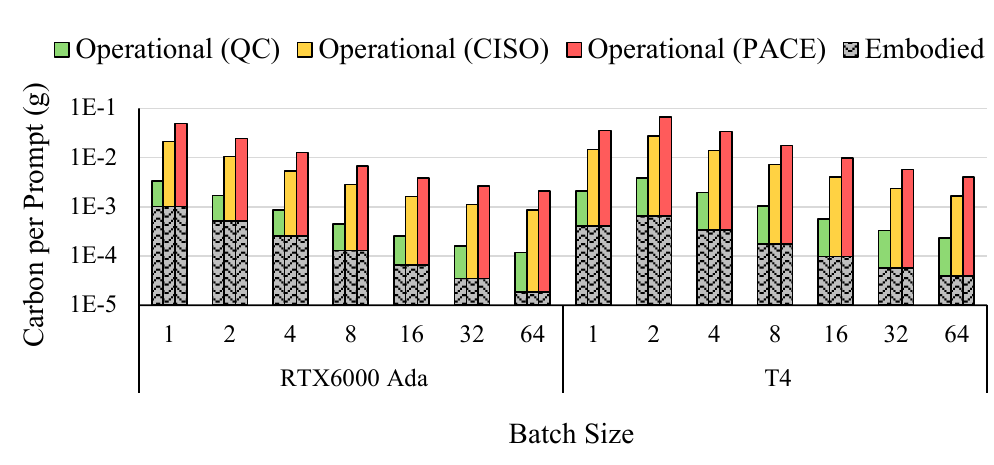}
    \caption{Per-prompt carbon emission under the QC, CISO, and PACE grids (1B-parameter LLaMA).}
    \label{fig:carbon_location}
    \Description{}
    \vspace{-0.3cm}
\end{figure}

\Cref{fig:carbon_location} shows the per-prompt operational and embodied carbon emissions (log-scaled y-axis) with different batch sizes (x-axis) in the three regions of \Cref{tab:carbon_intensity}: QC, CISO, and PACE.

\Cref{subsec:lat_eng} demonstrates that the newer \ada GPU is more energy-efficient than the older \tf except when the batch size is 1. 
When considering CI, the same conclusion remains true within the same region.
However, when comparing among regions, the older \tf may yield a lower operational carbon emission if it is located in a region with a relatively lower CI.
For instance, \tf in QC leads to lower operational carbon emissions compared to \ada in CISO or PACE, even when the batch size is 64.
Especially after factoring in the embodied carbon of \tf, the total carbon emissions of \tf are lower than those of \ada in low-CI regions. 

We observe that the embodied carbon, which is the bottom stack in the log-scaled bar chart, is a relatively small fraction of total carbon emissions across all regions.
According to \Cref{equ:em}, the embodied carbon is independent of CI because it is determined at manufacturing time.
In contrast, the operational carbon varies and depends on the grid carbon intensity, according to \Cref{equ:op}.
Therefore, the operational carbon is lower when the same prompt is executed in a region with a lower CI. 
For \tf, the embodied carbon comprises up to 19.7\%, 2.8\%, and 1.2\% of the total carbon emissions in QC, CISO, and PACE, respectively.
For \ada, the embodied carbon comprises up to 30.7\%, 5.0\%, and 2.1\% in the same regions. 
This disparity arises from \ada's newer CMOS technology, larger chip area, and higher onboard memory capacity, resulting in significantly higher embodied carbon.
These results indicate that embodied carbon emissions carry greater weight when GPUs are powered by grids with lower CIs.


\takeawaybox{
The balance between operational and embodied carbon emissions varies, depending on the CI. In high-CI regions, operational carbon makes up a significant portion of per-prompt carbon emissions, while in low-CI regions, embodied carbon becomes more prominent. Consequently, older and less powerful GPUs may be preferable in low-CI regions, whereas newer and more powerful GPUs are preferable in high-CI regions.

}

\subsection{Carbon Emissions in Prefill/Decode Phases}
\label{subsec:prefill_decode_carbon}
\begin{figure}
\begin{subfigure}[b]{0.48\linewidth}
    \includegraphics[height=3.5cm]{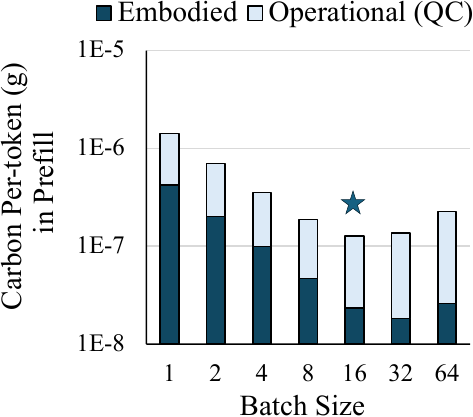}
    \caption{\ada}\label{subfig:1b_prefill_carbon_ada6000}
\end{subfigure}
\hspace{1pt}
\begin{subfigure}[b]{0.48\linewidth}
    \includegraphics[height=3.5cm]{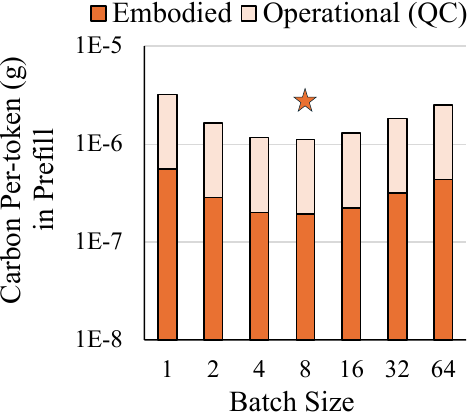}
    \caption{\tf}\label{subfig:1b_prefill_carbon_t4}
    \Description{}
\end{subfigure}
\caption{Per-token carbon emission in the \emph{prefill} phase under the CISO grid (1B parameters).}
\label{fig:1b_prefill_carbon}
\Description{}
\end{figure}

\begin{figure}
\begin{subfigure}[b]{0.48\linewidth}
    \includegraphics[height=3.5cm]{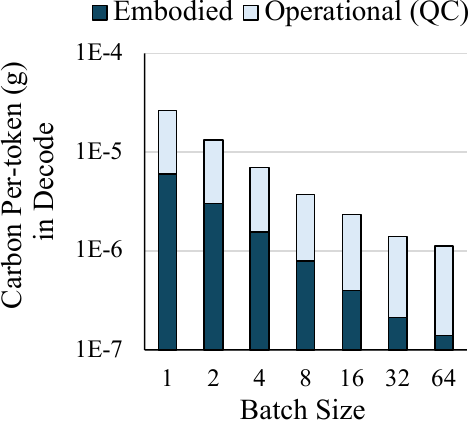}
    \caption{\ada}\label{subfig:1b_decode_carbon_ada6000}
    \Description{}
\end{subfigure}
\hspace{1pt}
\begin{subfigure}[b]{0.48\linewidth}
    \includegraphics[height=3.5cm]{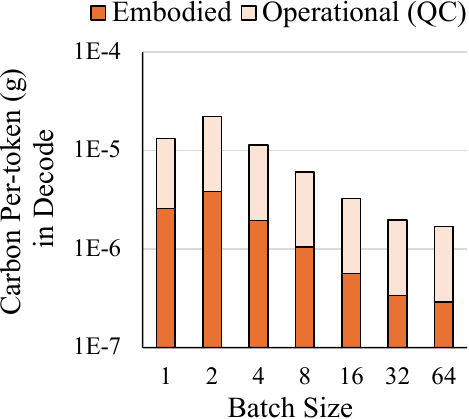}
    \caption{\tf}\label{subfig:1b_decode_carbon_t4}
    \Description{}
\end{subfigure}
\caption{Per-token carbon emission in the \emph{decode} phase under the CISO grid (1B-parameter LLaMA).}
\label{fig:1b_decode_carbon}
\Description{}
\end{figure}

We further study the operational and embodied carbon emissions in the prefill and decode phases of the 1B parameter version.
We use the QC's CI value to calculate the operational carbon emissions. 

\Cref{fig:1b_prefill_carbon} depicts the per-token carbon emissions (g) in the prefill phase that consists of embodied (bottom stack) and operational (upper stack) carbon emissions, respectively. Overall, the result is similar to the per-token energy numbers in \Cref{subfig:1b_prefill_energy}, as the operational carbon which dominates the total carbon emission is proportional to the energy consumption. 
However, with the inclusion of embodied carbon, we observe that the trend is not exactly the same.
A batch size of 16 consumes the lowest energy in \Cref{subfig:1b_prefill_energy}, 14.0\% lower than the second-best configuration which is the batch size of 32.
In comparison, when it comes to the total carbon emissions, the difference between batch sizes of 16 and 32 becomes smaller -- the former is only 7.3\% lower than the latter.
Similarly, \Cref{fig:1b_decode_carbon} shows the per-token carbon emissions (g) in the decode phase. The result also follows the same trend as the per-token energy numbers in \Cref{subfig:1b_decode_energy} with a smaller difference between \tf and \ada due to \tf's lower embodied carbon emissions.
We expect that for regions with lower CIs than QC (e.g., those powered by 100\% renewable energy), older GPUs like \tf could potentially yield even lower total carbon emissions, provided latency requirements are met, owing to their lower embodied carbon.


\takeawaybox{ Energy efficiency is not equivalent to carbon efficiency. To quantify environmental impacts, carbon-based metrics are more suitable than energy-based metrics. Factoring in embodied carbon emissions yields different optimal configurations compared to those that do not. Simultaneously accounting for both embodied and operational carbon reveals more optimization opportunities than considering either alone.
}

\subsection{Impact of Extending GPU Lifetime}
\label{subsec:lifetime}

\begin{figure}
    \centering
    \includegraphics[width=1\linewidth]{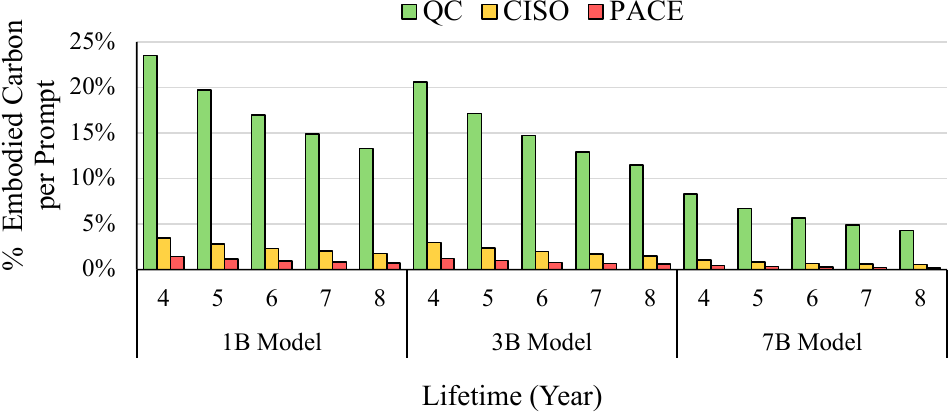}
    \vspace{-0.3cm}
    \caption{Embodied carbon emissions of \tf GPU under different expected lifetimes (batch size of 1).}
    \label{fig:carbon_year}
    \Description{}
\end{figure}

The previous analysis assumes a GPU lifespan of 5 years, which is typical for datacenter server components, as discussed in \Cref{subsec:carbon_meth}.
\Cref{fig:carbon_year} depicts the anticipated percentage of embodied carbon emissions over the total per-token carbon emissions (y-axis) by sweeping the lifetime of the older \tf GPU from 4 to 8 years (x-axis) in different regions. 
We observe that a reduced lifetime increases the percentage of embodied carbon emissions, while an increased lifetime reduces this percentage, as it is amortized over a longer time span. 
In regions with lower CIs, this increase is more prominent as embodied carbon emissions occupy a large fraction of the total carbon emissions. 
The same trend holds true for the 1B, 3B, and 7B parameter models, except that the embodied carbon emissions take up a lower percentage in larger models, as they are more compute-intensive and lead to more operational carbon emissions. 


\takeawaybox{
Extending the lifetime of GPUs results in lower embodied carbon emissions, as they are amortized over a longer duration. This decrease in embodied carbon emissions from using older GPUs is particularly prominent in regions with lower CIs.

}

\section{Future Directions } \label{sec:future}


In this section, we discuss future directions based on the findings and insights from this work. 

\textbf{Designing ML software-hardware infrastructure with long-term sustainability.}
Latest-generation GPUs and ML accelerators are often in shortage. 
Our study has shown that old GPUs can be usable in many scenarios and may lead to even lower total carbon emissions, making old hardware reuse a viable approach to bridge the gap between the demand and limited supplies of new GPUs. 
When designing hardware infrastructure for LLMs, it is critical to keep the whole lifetime in consideration, rather than optimizing for the current, specific applications. 
For example, the memory capacity can be a deal-breaker -- latest LLMs do not fit into older GPUs with small onboard memory; an awareness of interconnect and extensibility can help improve the lifespan of these ML accelerators. 
On the other hand, to better leverage old GPUs or ML accelerators, we envision optimizations for old platforms throughout the LLM software stack, from LLM algorithms, to runtimes, to the ML compiler level.   
Together, we expect efforts in these directions to pave the way to achieve longevity in ML infrastructure.

\textbf{Characterization of diverse LLM hardware platforms.}
This work focuses on GPUs, the dominant processors for LLM.
As demand for ML infrastructure rises, particularly with the widespread adoption of LLMs, datacenters are increasingly integrating specialized ML accelerators, such as TPUs in Google cloud~\cite{google_tpu}, MTIA accelerators in Meta cloud~\cite{meta_mtia}, and Trainum in AWS~\cite{aws_trainium}. 
These ML accelerators possess distinct performance, energy consumption, and embodied carbon compositions compared to GPUs.
Therefore, fully understanding the performance and carbon emission tradeoffs of specialized ML accelerators is critical to achieving the sustainability of LLM serving in the clouds.
Moving forward, we anticipate the development of profiling frameworks tailored to assess the carbon emissions of diverse GPUs and ML accelerators.

\textbf{CI-directed LLM serving.}
Our study has shown that the operational carbon emissions of LLM serving largely depend on the CI.
When the CI is high, the operational carbon emissions dominate the total carbon emissions.
Conversely, in cases of low CI, the embodied carbon emissions have a higher weight, making the use of older GPUs more beneficial. 
Prior work has demonstrated significant variability in carbon intensity across both temporal and spatial scales, as renewable energy sources are intermittent~\cite{maji2022carboncast,maji2023bringing,zhang2023gnn}. The flexibility in workload execution and carbon intensities presents opportunities for datacenter providers and users to schedule LLM workloads on GPUs that best match the current carbon intensity levels.
Further, CI predictions \cite{maji2022carboncast,maji2022dacf} can work collaboratively with the CI-directed scheduling strategy to make early scheduling decisions for LLM workloads and infrastructure.

\textbf{Sustainable LLM training.} 
While our focus lies on LLM serving, there is considerable potential for sustainable LLM training. Training LLMs is highly compute-demanding. 
For instance, Google trained their 540B-parameter PaLM model using 6144 TPU v4's\cite{chowdhery2022palm}. 
Hence, there's a pressing need to reduce the massive operational carbon emissions of LLM training. Two directions can be explored. First, unlike latency-critical LLM serving, LLM training offers flexibility as it lacks strict deadlines, allowing for easy workload shifting to periods and regions with lower carbon intensity. Second, in fine-tuning scenarios that are less compute-intensive~\cite{eliad2021finetuning,ye2023aspen}, older and lower-end GPUs/accelerators can be better utilized for such tasks.

\section{Conclusions}\label{sec:conclusion}

Serving LLMs on a large scale results in significant environmental impacts. 
This work studies both operational and embodied carbon emissions of LLaMA across various model parameter sizes and batch sizes using two Nvidia GPU types: \ada and \tf. 
Our findings indicate several optimization opportunities for reducing both operational and embodied carbon emissions for LLM serving. We hope this work inspires system researchers and developers to consider environmental impacts for building next-generation sustainable LLM serving systems. 


\begin{acks}
We thank the anonymous reviewers for their valuable feedback, and Tianyao Shi and Jiashu Zhang for proofreading. 
We acknowledge GCP and AWS for providing generous research credits.
This work is supported by the Natural Sciences and Engineering Research Council of Canada (NSERC) and the Undergraduate Research Assistantship (URA) program of the Cheriton School of Computer Science at the University of Waterloo.
\end{acks}

\bibliographystyle{plain}
\bibliography{reference}

\end{document}